# Classical Chinese Sentence Segmentation for Tomb Biographies of Tang Dynasty

Chao-Lin Liu and Yi Chang
National Chengchi University, Taiwan
chaolin@nccu.edu.tw, black.heptagram@gmail.com

**Introduction**

Figure 1 shows a slab of tomb biography of the Tang dynasty [1]. Researchers can copy the words on such slabs to produce a collection of tomb biographies for research. A typical tomb biography contains various types of information about the deceased and their families and, sometimes, a rhyming passage of admiration. Employing software tools to analyze the texts, one can extract useful information from the collections of tomb biographies to enrich databases like the China Biographical Database (CBDB)[2] to support further Chinese studies.

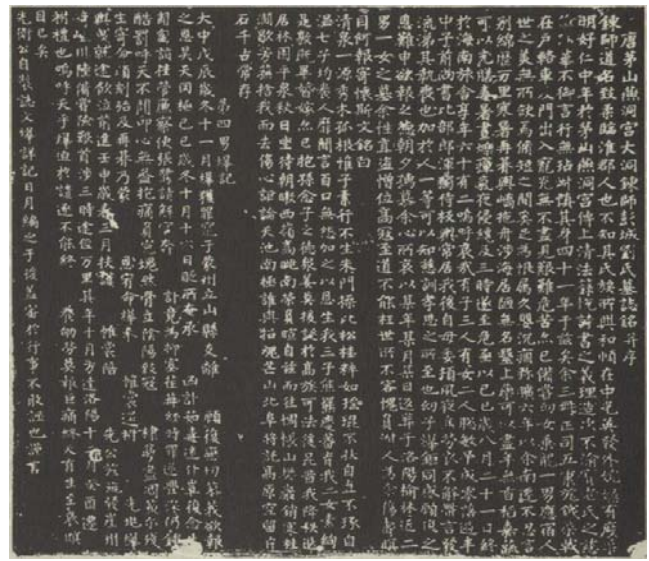

**Figure 1. A slab of tomb biography of the Tang dynasty**

It is well known that modern Chinese texts do not include delimiters like spaces to separate words. Hence, researchers design algorithms for segmenting Chinese character strings into words[3,4].

In contrast, it is not as well known that, in classical Chinese, there were no markers for the separation of sentences. The characters in Figure 1 simply connect to each other. In modern Chinese, texts are punctuated for pauses in sentences and ends of sentences. The research about algorithmically inserting these syntactic markers into classical Chinese is receiving more attention along with the growth of digital humanities in recent years. The needs of segmenting ancient texts for humanities studies are not unique to Chinese studies, interested readers can find examples for German texts[5] and Swedish texts[6].

---

[1] This image was downloaded from <http://www.lyqtzz.com/uploadfile/20110817165325665.jpg>. The Tang dynasty existed between 688CE and 907CE. More images of tomb biographies are available at <http://goo.gl/XHCL9P>.
[2] The China Biographical Database (https://projects.iq.harvard.edu/cbdb/home) is a free and open database for Chinese studies.
[3] Mao-Song Sun, Min Xiao, and B. K. Tsou. 2004. Chinese word segmentation without using dictionary based on unsupervised learning strategy, *Chinese Journal of Computers*, 27(6):736–742. (in Chinese)
[4] Yan Shao, Christian Hardmeier, Jörg Tiedemann, and Joakim Nivre. 2017. Character-based joint segmentation and POS tagging for Chinese using bidirectional RNN-CRF, *Proceedings of the 2017 International Joint Conference on Natural Language Processing*.
[5] Florian Petran. 2012. Studies for segmentation of historical texts: Sentences or chunks? *Proceedings of the Second Workshop on Annotation of Corpora for Research in the Humanities*, 75–86.
[6] Gerlof Bouma and Yvonne Adesam. 2013. Experiments on sentence segmentation in Old Swedish editions. *Proceedings of the Workshop on Computational Historical Linguistics at NODALIDA 2013*. NEALT Proceedings Series 18 / Linköping Electronic Conference Proceedings 87: 11–26.

Huang et al.[7] employed the techniques of conditional random fields (CRFs) to segment texts of literature and history. They achieved 0.7899 and 0.9179 in $F_1$[8], respectively, for segmenting the texts in Shiji and Zuozhuan[9]. Wang et al.[10] applied recurrent neural networks to segment texts in a diverse collection of classical Chinese sources. They achieved $F_1$ measures that are close to 0.75, and item accuracies[11] that are near 0.91. The researchers achieved different segmentation results for different corpora even when they applied the same techniques and procedures. It is thus inappropriate to just compare the numbers for ranking because the nature of the corpora varies widely.

In this proposal, we report our attempts to segment texts in tomb biographies with CRF models[12]. We studied the effects of considering different types of lexical information in the models, and achieved 0.853 in precision[8], 0.807 in recall, 0.829 in $F_1$, and 0.940 in item accuracy[13].

**Data Sources**

We obtained digitized texts for three books of tomb biographies of the Tang dynasty[14]. The collection consists of 5119 biographies which contain 423,922 periods, commas, and semicolons. There are 5505 distinct types of characters and a total of more than 2461 thousand of characters in the collection[15]. When counting these statistics, we ignored a very small portion of characters that cannot be shown without special fonts. Hence, these statistics are not perfectly precise, but they are accurate within a reasonable range. On average, a biography has about 480 characters. Some of them are very short and have many missing characters. Hence, we exclude biographies that have no more than 30 characters[16] in our experiments.

---

[7] Hen-Hsen Huang, Chuen-Tsai Sun, and Hsin-Hsi Chen. 2010. Classical Chinese sentence segmentation, *Proceedings of CIPS-SIGHAN Joint Conference on Chinese Language Processing*, 15–22.

[8] The **precision** rate, **recall** rate, and **F measure** are designed for evaluating the effectiveness of information retrieval and extraction. $F_1$ is a popular choice of the F measure.

[9] Shiji (史記) and Zuozhuan (左傳) are two very important sources about Chinese history.

[10] Both Chinese and English versions are available.
- Boli Wang, Xiaodong Shi, and Jinsong Su. 2017. A sentence segmentation method for ancient Chinese texts based on recurrent neural network, *Acta Scientiarum Naturalium Universitatis Pekinensis*, 53(2):255–261. (in Chinese)
- Boli Wang, Xiaodong Shi, Zhixing Tan, Yidong Chen, and Weili Wang. 2016. A Sentence Segmentation Method for Ancient Chinese Texts Based on NNLM. *Proceedings of the Chinese Lexical Semantics Workshop 2016*, Lecture Notes in Computer Science 10085, 387–396.

[11] The **item accuracy** evaluates the labeling judgments including both punctuated and non-punctuated items. In a typical sentence segmentation task, there are many more non-punctuated items than punctuated items, so it is relatively easier to achieve attractive figures for the item accuracy than for the F measure.

[12] John Lafferty, Andrew McCallum, and Fernando C.N. Pereira. 2001. Conditional random fields: Probabilistic models for segmenting and labeling sequence data, *Proceedings of the Eighteenth International Conference on Machine Learning*, 282–289.

[13] We interviewed Hongsu Wang (王宏甦), the project manager of the China Biographical Database Project at Harvard University, about his preferences in post-checking the segmentation results that are produced by software. He suggests that higher precision rates are preferred. When seeking higher recall rates (often sacrificing the precision rates), the false-positive recommendations for punctuation are annoying to the researchers.

[14] We obtained the texts from the following sources.
(1) Shaoliang Zhou (周紹良) and Chao Zhao (趙超). 1992. *A Collection of Tomb Biographies of Tang Dynasty* (唐代墓誌彙編), Shanghai Ancient Books Publishing House (上海古籍出版社). (in Chinese)
(2) Shaoliang Zhou (周紹良) and Chao Zhao (趙超). 2001. *A Collection of Tomb Biographies of Tang Dynasty*: *An Extension* (唐代墓誌彙編續集), Shanghai Ancient Books Publishing House. (in Chinese)

[15] In terms of Linguistics, we have 5505 character types and 2,461,000 character tokens.

**Training and Testing CRF Models**[17]

We consider the segmentation task as a classification problem. Let $C_i$ denote an individual character in the texts. We categorize each character as either **M** (for "followed by a punctuation mark") and **O** (for "an ordinary character"). Assume that we should add only a punctuation mark between $C_3$ and $C_4$ for a string "$C_1$ $C_2$ $C_3$ $C_4$ $C_5$". A correct labeling for this string will be "O O M O O".

We can convert each character in the texts into an *instance*, which may be used for training or testing. We provide with each instance a group of contextual *features* that may be relevant to the judgment of whether or not a punctuation mark is needed. For instance, we may use one character surrounding a character X and itself as the group of features. The following are two instances that we create for $C_3$ and $C_4$. The instance for $C_3$ is (1), and the leftmost item is the correct label for $C_3$, and the rest are the features[18] for $C_3$.

$$M \quad w[0]=C_3, w[-1]=C_2, w[1]=C_4 \quad (1)$$
$$O \quad w[0]=C_4, w[-1]=C_3, w[1]=C_5 \quad (2)$$

We can train a CRF model with a selected portion of the instances (called *training data*), and test the resulting model with the remaining instances (called *test data*). The instances in the training and the test data are mutually exclusive.

We employ a machine-learning tool[19] that learns from the training data to build a CRF model. We then apply the learned model to predict the classes of the instances in the test data. The labels of the instances in the test data are temporarily concealed[20] when the learned model makes predictions. The precision rate and recall rate of the learned model are calculated with the correct and the predicted labels.

We report four sets of basic experiments next, each investigating an important aspect for analyzing Chinese texts. The 5119 biographies[21] were resampled and randomly assigned to the training (70%) and test (30%) sets for every experiment. We repeated every experiment three times, and report the averages of the precision and recall rates.

---

[16] 30 is an arbitrary choice, and can be changed easily.
[17] Due to the constraint on the word count in DH 2018 proposals, we can only briefly outline the steps for training and testing CRF models. More details can be provided in the presentation and in an extended report.
[18] Here, we adopt typical notations for CRF-based applications. w[0] is the current word, w[-1] is the neighbor word to the left of the current word, w[1] is the neighbor word to the right of the current word. Two actual instances that are produced from "孝敬天啟，動必以禮" for character-based segmentations will look like the following.
  O   w[-1]=敬,w[0]=天,w[1]=啟
  M   w[-1]=天,w[0]=啟,w[1]=動
Two instances that are produced from "母子 忠孝 ， 天下 榮 之" for the word-based segmentations will look like the following.
  M   w[-1]=母子,w[0]=忠孝,w[1]=天下
  O   w[-1]=忠孝,w[0]=天下,w[1]=榮
[19] CRFSuite: <http://www.chokkan.org/software/crfsuite/>
[20] Thus, the instances for testing CRFs look like (1) and (2) that do not carry the correct labels, M and O, respectively..
[21] Recall that we used only those biographies that have no less than 30 characters.

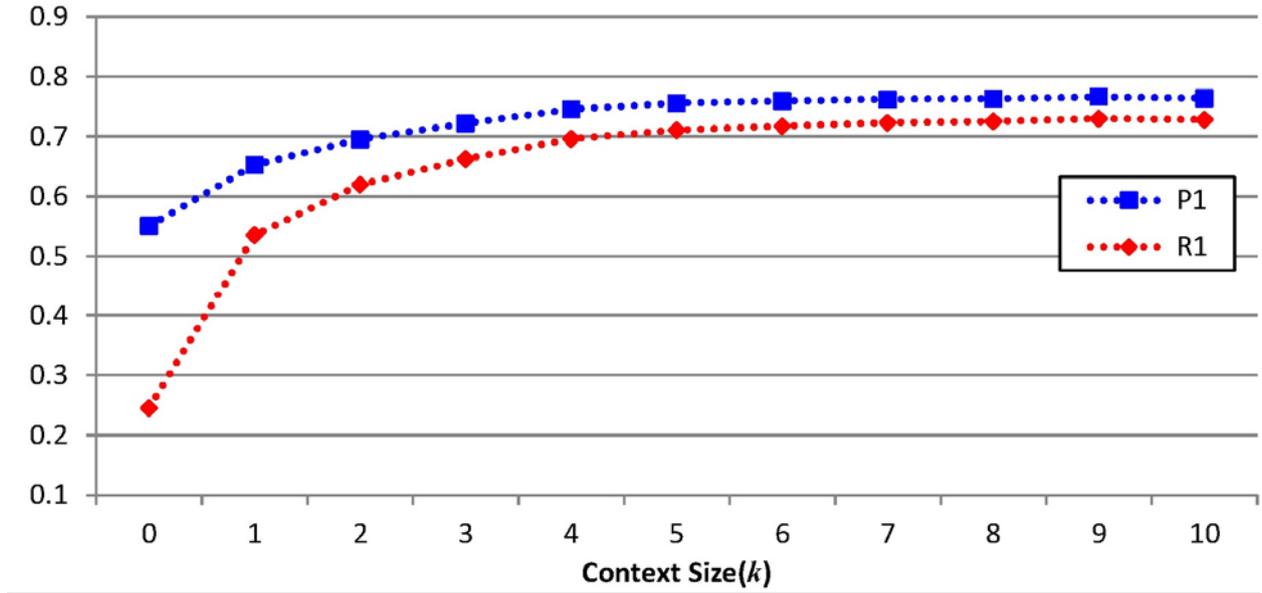

Figure 2. Effects of varying context sizes

### Changing the Size of the Context

We certainly can and should consider more than one character around the current character as the context. Figure 2 shows the test results of using different sizes of contexts for the instances. The horizontal axis shows the sizes, e.g., when $k$=2, the feature set includes information about two characters on both sides of the current character. P1 and R1 are the average precision and recall rates, respectively.

We expected to improve the precision and recall rates by expanding the width of the context. The margin of improvements gradually decreased, and the curves level off after the window sizes reached six. The recall rises sharply when we add the immediate neighbor word into the features, emphasizing the predicting power of the immediate neighbor character. When $k$=10, the precision and recall are 0.765 and 0.729, respectively, and the item accuracy exceeds 0.91.

### Adding Bigrams

We added bigrams that were formed by consecutive characters into the features. The following instance shows the result of adding bigrams to the features in (1)[22].

$$M \quad w[0]=C_3, w[-1]=C_2, w[1]=C_4, \mathbf{w[-1\_0]=C_2C_3, w[0\_1]=C_3C_4} \quad (3)$$

Figure 3 shows the test results of adding bigrams while we also tried different sizes of context. The curves named P1 and R1 are from Figure 2, and P2 and R2 are results achieved by adding bigrams to the features. Both rates are improved, and the gains are remarkable.

---

[22] Here, w[-1_0] is the bigram on the left side of the current word, and w[0_1] is the bigram to the right of the current word. When we consider bigrams for a wider context, we may consider bigrams like w[-2_-1] and w[1_2].

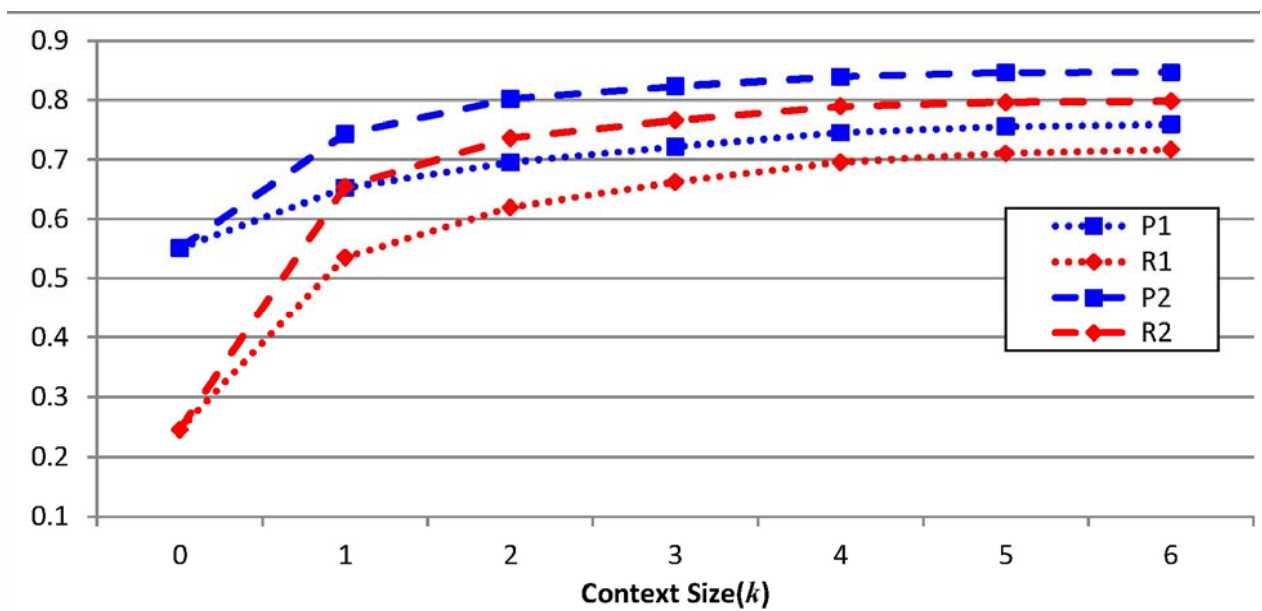

**Figure 3. Adding bigrams improves the results.**

**Effects of Pronunciation Information**

Using the characters and their bigrams in the features is an obvious requirement. Since the tomb biographies may contain rhyming parts, it is also intriguing to investigate whether adding pronunciation information may improve the overall quality of the segmentation task.

We considered two major sources of the pronunciation information for Chinese characters in the Tang dynasty: *Guangyun* and *Pingshuiyun*[23]. The statistics in Table 1 show that adding pronunciation information into the features did not improve the overall performance for the segmentation task significantly[24]. The results suggest that, given the characters and their bigrams, adding pronunciation did not contribute much more information. Huang et al.[7] reported similar observations when they used *Guangyun* in their work. Relatively, *Guangyun* is more informative than *Pingshuiyun* for the segmentation tasks.

**Adding Word-Level Information**

We can obtain information about the reign periods, location names, and office names in the Tang dynasty from CBDB. By segmenting characters for these special words and adding appropriate type information, we added word-level information into the features. The statistics in Table 2 show that the word-level information[25] did not raise the performance very much.

We examined the training and test data, and found that, although we gathered the special terms for the Tang dynasty, those words were not used in the biographies often. As a consequence, we did not add a lot of word-level information in the features in reality.

---

[23] *Guangyun* and *Pingshuiyun* are 《廣韻》 and 《平水韻》, respectively
[24] This does not suggest that using the pronunciation information alone was not useful. We have conducted more experiments to evaluate the effectiveness of using the pronunciation information for the segmentation tasks, and will provide more details in the presentation and in an extended report.
[25] In Table 2, WOC stands for "Width of Context", "P" stands for precision, "R" stands for recall, "C+B" stand for "Characters and Bigrams" and "C+B+W" stands for "Characters, Bigrams, and Words".

| Features | Width of Context = 1 | | | Width of Context = 2 | | |
|---|---|---|---|---|---|---|
| | Precision | Recall | $F_1$ | Precision | Recall | $F_1$ |
| Characters | 0.652 | 0.535 | 0.588 | 0.695 | 0.620 | 0.655 |
| Characters+Bigrams | 0.743 | 0.654 | 0.696 | 0.802 | 0.736 | 0.768 |
| Characters+Bigrams+Guangyun | 0.748 | 0.671 | 0.707 | 0.781 | 0.707 | 0.742 |
| Characters+Bigrams+Pingshuiyun | 0.737 | 0.659 | 0.696 | 0.763 | 0.698 | 0.729 |

**Table 1. Contributions of pronunciation information**

| Features | WOC = 1 | | WOC = 2 | | WOC = 3 | | WOC = 4 | | |
|---|---|---|---|---|---|---|---|---|---|
| | P | R | P | R | P | R | P | R | $F_1$ |
| C+B | 0.743 | 0.654 | 0.802 | 0.736 | 0.823 | 0.766 | 0.839 | 0.790 | 0.814 |
| C+B+W | 0.747 | 0.671 | 0.800 | 0.741 | 0.818 | 0.767 | 0.832 | 0.787 | 0.809 |
| C+B+PMI | 0.748 | 0.661 | 0.804 | 0.740 | 0.824 | 0.769 | 0.839 | 0.791 | 0.814 |

**Table 2. Adding word-level information**

We have also adopted pointwise mutual information (PMI) of bigrams as features, but the net contributions are not significant.

**Discussions**

We have consulted historians[13,26], and learned that our current results are useful in practice. The best precision rates and F measures are better than 0.8 in Figure 3 and Table 4. The best item accuracy is better than 0.94.

In fact, we have designed an advanced mechanism to further improve our results[27]. The new approach employs a second level learning step that learns from the errors of the current classifiers.

One may plan to consider more linguistic information in the segmentation tasks. If appropriate corpora or sources are available, it is worthwhile to explore the effects of adding part-of-speech information in the task[28,29].

Although we look for methods to reproduce the segmentations in the given texts, we understand that not all experts will agree upon "the" segmentations for a corpus. Different segmentations may correspond to different interpretations of the texts, especially for the classical Chinese. The results of asking two persons to segment Chinese texts may not match perfectly either[30].

---

[26] In addition to Hongsu Wang of Harvard University, we also consulted Professor Zhaoquan He (何兆泉) of the China Jiliang University. They use tomb biographies of the Tang and the Song dynasties in their research.

[27] Again, we could not provide details about more experiments because of the word limit for DH 2018 submissions.

[28] Tin-shing Chiu, Qin Lu, Jian Xu, Dan Xiong, and Fengju Lo. 2015. PoS tagging for classical Chinese text, *Chinese Lexical Semantics* (Lecture Notes in Artificial Intelligence 9332), 448–456.

[29] John Lee. 2012. A classical Chinese corpus with nested part-of-speech tags, *Proceedings of the Sixth EACL Workshop on Language Technology for Cultural Heritage, Social Sciences, and Humanities*, 75–84.

[30] Hen-Hsen Huang and Hsin-Hsi Chen. 2011. Pause and stop labeling for Chinese sentence boundary detection, *Proceedings of the 2011 Conference on Recent Advances in Natural Language Processing*, 146–153.